\documentclass[sigconf, screen, final]{acmart}

\newcommand{\systemname}{PG\allowbreak OV\allowbreak 3D}
\newcommand{\consistencymodule}{inter-frame \allowbreak consistency \allowbreak module}
\newcommand{\llavanext}{LLaVA-NeXT}

\AtBeginDocument{%
  }

\usepackage{multirow} 
\usepackage{colortbl} 

\usepackage{booktabs}
\usepackage{multirow}
\usepackage{graphicx}
\usepackage{bbding}
\usepackage{enumitem}
\usepackage{pifont}

\setcopyright{acmlicensed}
\copyrightyear{2018}
\acmYear{2018}
\acmDOI{XXXXXXX.XXXXXXX}
\acmConference[Conference acronym 'XX]{Make sure to enter the correct
  conference title from your rights confirmation email}{June 03--05,
  2025}{Woodstock, NY}
\acmISBN{978-1-4503-XXXX-X/2025/06}

\acmSubmissionID{5650}
\begin{document}


\title{PGOV3D: Open-Vocabulary 3D Semantic Segmentation with Partial-to-Global Curriculum}


\author{Shiqi Zhang}
\email{zhangshiqi_1127@mail.ustc.edu.cn}
\affiliation{
  \institution{University of Science and Technology of China}
  \city{Hefei}
  \country{China}}

\author{Sha Zhang}
\email{zhsh1@mail.ustc.edu.cn}
\affiliation{
  \institution{University of Science and Technology of China}
  \city{Hefei}
  \country{China}}

\author{Jiajun Deng}
\email{jiajun.deng@adelaide.edu.au}
\authornote{Corresponding author}
\affiliation{
\institution{The University of Adelaide}
\city{Adelaide}
\country{Austrilia}
}

\author{Yedong Shen}
\email{sydong2002@mail.ustc.edu.cn}
\affiliation{
  \institution{University of Science and Technology of China}
  \city{Hefei}
  \country{China}}

\author{Mingxiao Ma}
\email{mingxiaoma@ustc.edu.cn}
\affiliation{
  \institution{University of Science and Technology of China}
  \city{Hefei}
  \country{China}}

\author{Yanyong Zhang}
\authornote{Corresponding author}
\email{yanyongz@ustc.edu.cn}
\affiliation{
  \institution{University of Science and Technology of China}
  \city{Hefei}
  \country{China}}



%
\begin{abstract}

Existing open-vocabulary 3D semantic segmentation methods typically supervise 3D segmentation model by merging text-aligned features (e.g., CLIP) extracted from multi-view images onto 3D points. However, such approaches treat multi-view images merely as intermediaries for transferring open-vocabulary information, overlooking their rich semantic content and cross-view correspondences, and thus limiting the model's effectiveness. To this end, we propose \textbf{\systemname}, a novel framework that introduces \underline{\textbf{P}}artial-to-\underline{\textbf{G}}lobal curriculum to improve \underline{\textbf{O}}pen-\underline{\textbf{V}}ocabulary \underline{\textbf{3D}} semantic segmentation. 
The key innovation of our work is a two-stage training strategy. \textbf{In the first stage}, we pre-train the model on partial scenes that provide dense semantic information but relatively simple geometry. 
Partial point clouds are derived from multi-view RGB-D inputs via pixel-wise depth projection. To enable open-vocabulary learning, we leverage a multi-modality large language model (MLLM) and a 2D segmentation foundation model to generate open-vocabulary labels for each viewpoint, providing rich and aligned supervision. An auxiliary \consistencymodule\ is introduced during this stage to enforce feature consistency under viewpoint variations and enhance spatial understanding. \textbf{In the second stage}, we fine-tune the model on complete scene-level point clouds, which are sparser and structurally more complex. To support this, we aggregate the partial vocabularies associated with each scene and generate pseudo labels using the pre-trained model, effectively bridging the semantic gap between dense partial observations and large-scale 3D environments.
Extensive experiments on ScanNet, ScanNet200 and S3DIS benchmarks demonstrate that \systemname, achieves competitive performance in open-vocabulary 3D semantic segmentation. The code will be released.

\begin{figure}[t]
    \centering
    \includegraphics[width=1.0\linewidth]{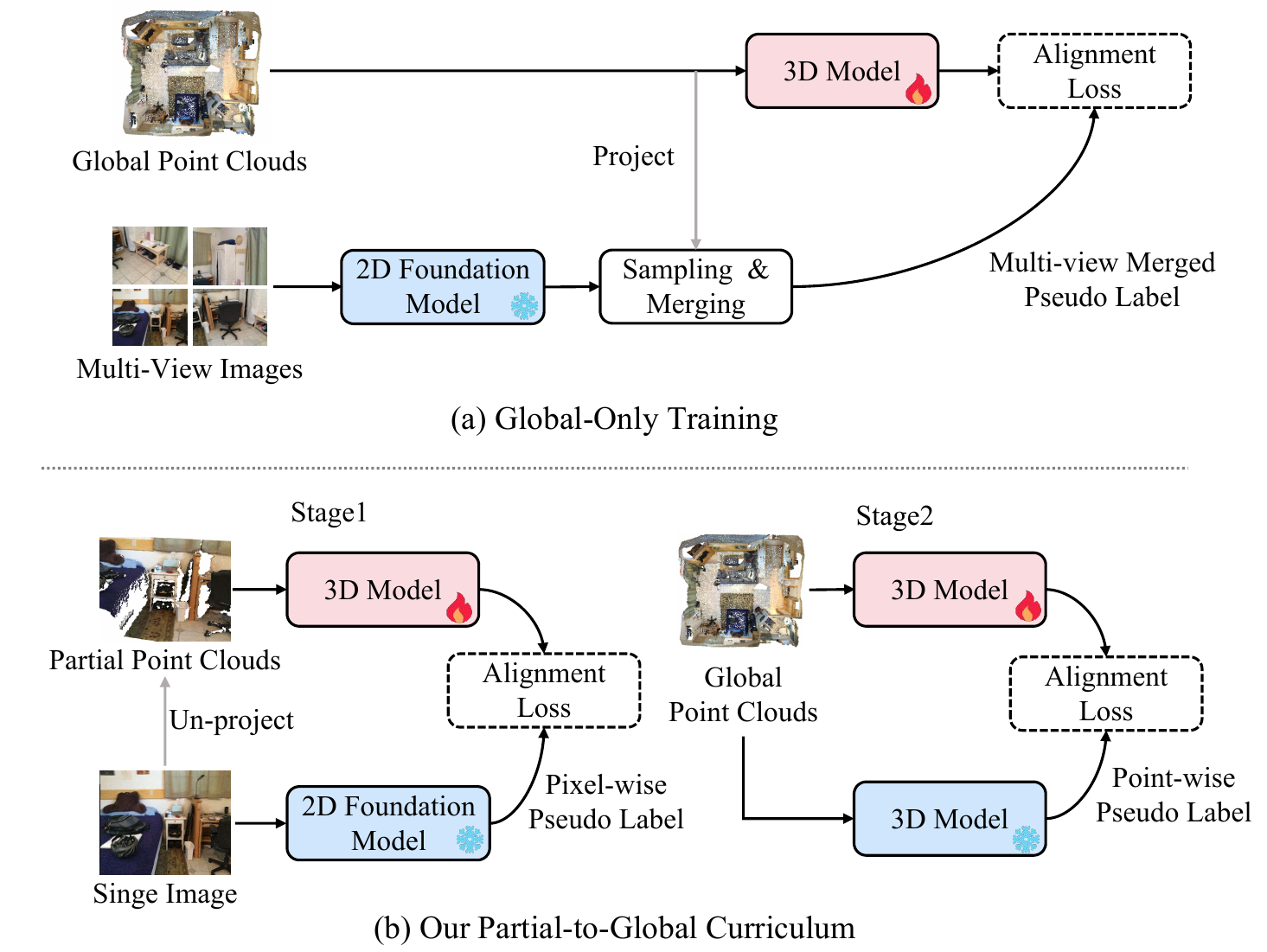} 
    \caption{\textbf{Global-Only Training VS Our Partial-to-Global Curriculum.} (a) The global-only training paradigm, sampling or merging features from multiple views onto 3D points, overlooks the rich semantic content in images. In contrast, (b) our proposed Partial-to-Global curriculum first supervise the 3D segmentation model via pixel-wise pseudo labels extracted by 2D foundation models, and subsequently fine-tune it with point-wise pseudo labels derived from the pre-trained 3D model. This two-stage training strategy exploit rich semantic content in images and cross-view correspondences.}
    \label{fig:intro}
\end{figure}

\end{abstract}





\maketitle

\section{Introduction}



3D semantic segmentation is fundamental to various real-world applications, such as autonomous driving and robotics, which involve complex environments and diverse object categories. To operate effectively in such open-world settings, models must generalize beyond a fixed set of predefined labels. However, conventional methods heavily rely on fully supervised learning with curated datasets, which are limited in both scale and category diversity. This hinders generalization to unseen objects and scenarios, and the high cost of dense 3D annotation further restricts scalability in practical deployments. As a result, open-vocabulary 3D semantic segmentation has emerged as a promising yet challenging direction, which attracts increasing research interest in the community.



The primary challenge of open-vocabulary 3D semantic segmentation lies in aligning 3D point clouds with textual semantics. However, due to the lack of large-scale point–text pairs, it is not feasible to directly apply contrastive learning as in the 2D domain (\emph{e.g.}, CLIP~\cite{radford2021learning}) to establish such an alignment. Alternatively, many existing approaches extract text-aligned features from multi-view images and transfer them to 3D points via projection~\cite{peng2023openscene, xiao20243d}, using the aggregated features as supervision for 3D semantic segmentation. Another line of work~\cite{ov3d2023, Ding_Yang_Xue_Zhang_Bai_Qi_2022} leverages large vision-language models to generate textual labels for images, which are then projected onto 3D points through geometric correspondences to obtain point-wise segmentation labels for point clouds.


Despite recent progress, existing approaches generally follow the global-only training paradigm, as shown in Figure~\ref{fig:intro} (a). This paradgim primarily leverage multi-view images as geometric bridges to transfer open-vocabulary image features or textual labels onto 3D points, while overlooking the rich semantic content present in the images themselves. Moreover, pixels corresponding to the same 3D point across different views often exhibit inconsistent features due to lighting variations, occlusions, and viewpoint shifts. As a result, merging or selecting features from multiple views can introduce ambiguity and additional noise.


To this end, we introduce \systemname, a novel framework built upon a Partial-to-Global curriculum for open-vocabulary 3D semantic segmentation illustrated in Figure \ref{fig:intro}(b). Partial scenes derived from RGB-D images offer high-resolution, pixel-level semantic details but limited geometric complexity, making them well-suited for learning localized features. In contrast, global point clouds capture richer contextual and structural information but are inherently sparse. To fully leverage the complementary strengths of these dual data types, our Partial-to-Global curriculum strategy first establishes point-semantic alignment using pixel-wise pseudo labels from RGB-D images, generated by pre-trained 2D foundation models. This is followed by refinement through point-wise supervision from the pre-trained 3D segmentation model. This two-stage training strategy leverages rich semantics in images and keep cross-view correspondences from multi-view sequences, effectively enhancing the performance of open-vocabulary 3D semantic segmentation.

Specifically, in the first stage, we convert the multi-view RGB-D sequences into dense partial point clouds through pixel-wise depth projection, enabling the preservation of fine-grained visual details. To generate open-vocabulary supervision, we utilize powerful MLLMs (\emph{e.g.}, LLaVA ~\cite{liu2024improved}, \llavanext ~\cite{liu2024llavanext}) and 2D open-vocabulary segmentation foundation model (\emph{e.g.}, SAM~\cite{kirillov2023segment}, SEEM~\cite{zou2023segment}) to produce pixel-wise semantics for each viewpoint as pseudo label. The 3D network aligns the feature space by training on partial point clouds paired with their corresponding pseudo labels.
However, due to occlusions and varying viewpoints, the same physical region may appear differently across frames, leading to semantic inconsistency among overlapping 3D points. Inspired by the proven effectiveness of multi-view consistency constraints in 3D visual reconstruction ~\cite{kerr2023lerf,qin2024langsplat}, we introduce an auxiliary \consistencymodule\ that explicitly enforces feature consistency across overlapping regions in different viewpoint. By encouraging the 3D network to maintain invariant feature representation for the same physical points across multi views, this module helps the model reconstruct more coherent 3D spatial representations from fragmented observations. However, a fundamental disparity persists between partial point clouds and their global counterparts. Therefore, in the second stage, we fine-tune the model on global point clouds, which are geometrically more complex and semantically diverse. We aggregate the partial vocabularies from different views within the same scene and high-confidence point-wise pseudo annotations are inferred through the predictions of the pre-trained segmentation model, serving as reliable supervision for fine-tuning phase. 

To validate the effectiveness of \systemname\ on open-vocabulary 3D semantic segmentation, we conduct extensive experiments on the ScanNet and ScanNet200 benchmarks. On ScanNet, our method achieves 59.5\% mIoU, improving the strongest competitor by a margin of 2.2\% mIoU, demonstrating strong zero-shot semantic segmentation capabilities. Furthermore, our model trained on ScanNet generalizes well to another dataset, S3DIS, without fine-tuning, demonstrating strong cross-domain generalization.

The main contributions of this paper are as follows:

\begin{itemize}[leftmargin=2em, topsep=0pt, itemsep=0.3em]
\item We introduce \systemname, a Partial-to-Global curriculum framework that first pre-trains on dense, simple partial scenes and then fine-tunes on sparse, complex global scenes. This approach addresses the issue of image content loss in the global-only training paradigm, while also facilitating seamless open-vocabulary feature alignment.
\item  We design a auxiliary \consistencymodule\ to encourage the 3D network maintaining invariant features for same physical points across multi views and help the model reconstruct more coherent 3D spatial representations from fragmented observations.
\item  We validate the proposed approach through extensive experiments on ScanNet, ScanNet200, and S3DIS. The results suggest that our method achieves competitive or superior zero-shot segmentation performance. 
\end{itemize}

\section{Related Work}

\subsection{Closed-set 3D Scene Understanding}

In recent years, substantial advances have been achieved in 3D scene understanding within closed datasets. This includes core tasks such as 3D object detection ~\cite{deng2024diff3detr, kolodiazhnyi2024oneformer3d, zhou2018voxelnet, pan20213d，su2020adapting}, semantic segmentation ~\cite{kirillov2023segment, wu2024point, zhao2021point}, and instance segmentation ~\cite{jiang2020pointgroup, lai2023mask, lu2023query, sun2023superpoint, vu2022softgroup, kolodiazhnyi2024oneformer3d}, all of which have seen notable improvements due to advances in neural architectures and large-scale annotated datasets. In parallel, recent works have begun to explore the integration of multimodal information—such as combining image and point cloud features—to further enhance 3D scene understanding within these closed-set frameworks.
Among the various tasks, semantic segmentation of point clouds plays a particularly central role, as it provides dense and fine-grained semantic information crucial for downstream applications such as autonomous driving and robotics. To accommodate diverse deployment scenarios, recent approaches have explored both online methods ~\cite{liu2022ins, zhang2020fusion}, which support real-time processing, and offline methods ~\cite{robert2022learning, schult2023mask3d, vu2022softgroup}, which typically achieve higher accuracy by leveraging full-scene context. Nevertheless, a fundamental limitation persists across both paradigms: they rely on a fixed, dataset-specific vocabulary defined during training. This constraint hinders their ability to generalize to novel categories and limits their applicability in open-world environments. 
\subsection{Open-Vocabulary 2D Scene Understanding}
With the rapid development of large-scale foundational language models ~\cite{chen2023clip2scene,cheng2024yolo,ding2023pla,li2022language,kirillov2023segment,liang2023open}, significant progress has been made in zero-shot 2D scene understanding. Building upon 2D foundational vision models, researchers have achieved breakthroughs in various tasks: works ~\cite{chen2023clip2scene,ding2023pla,peng2023openseene} proposed zero-shot 2D segmentation methods, studies ~\cite{cheng2024yolo,li2022language,liang2023open} explored open-vocabulary 2D image understanding, while ~\cite{ghiasi2022scaling,guo2024semantic,peng2023openseene} focused on open-vocabulary 2D object detection. 
Although numerous studies ~\cite{xu2023side,ghiasi2022scaling,li2022language,meng2023open,zou2023segment} have investigated open-vocabulary 2D segmentation, these methods typically require predicting open-vocabulary features at the pixel level by encoding 2D images and aligning them with open-vocabulary pixel features. However, due to the scarcity of large-scale 3D annotated data, directly transferring such approaches to end-to-end open-vocabulary 3D point cloud segmentation remains challenging.

\subsection{Open-Vocabulary 3D Scene Understanding}
Open-vocabulary 3D scene understanding has recently emerged as a promising research direction, with methods generally falling into two main paradigms based on their representational strategies. The first category of methods ~\cite{ov3d2023, peng2023openseene, wang2024open, zhu2024open} relies on depth cameras to acquire geometric information and establish 2D–3D correspondences through camera parameters. Representative works in this category include OpenScene~\cite{peng2023openseene}, which introduces a multi-view feature fusion strategy; OV3D~\cite{ov3d2023}, which leverages foundation models to enhance text-based supervision and improve feature alignment; GGSD~\cite{wang2024open}, which adopts a Mean Teacher framework for knowledge distillation; and Diff2scene~\cite{zhu2024open}, which utilizes text-to-image generative models for synthetic data augmentation.
The second category of approaches ~\cite{chen2023clip2scene, kerr2023lerf, qin2024langsplat, ye2024gaussian, shen2025og} distills 2D visual features into implicit representations, such as Neural Radiance Fields (NeRF) ~\cite{mildenhall2021nerf} or 3D Gaussian Splatting (3DGS) ~\cite{kerbl20233d}. Notable examples include LERF ~\cite{kerr2023lerf}, which builds on NeRF to incorporate language supervision, and LangSplat~\cite{qin2024langsplat}, which extends the 3DGS framework for open-vocabulary understanding in 3D scenes.
\section{Method}
\label{sec:method}
\begin{figure*}[thbp]
    \centering
    \includegraphics[width=1.0\linewidth]{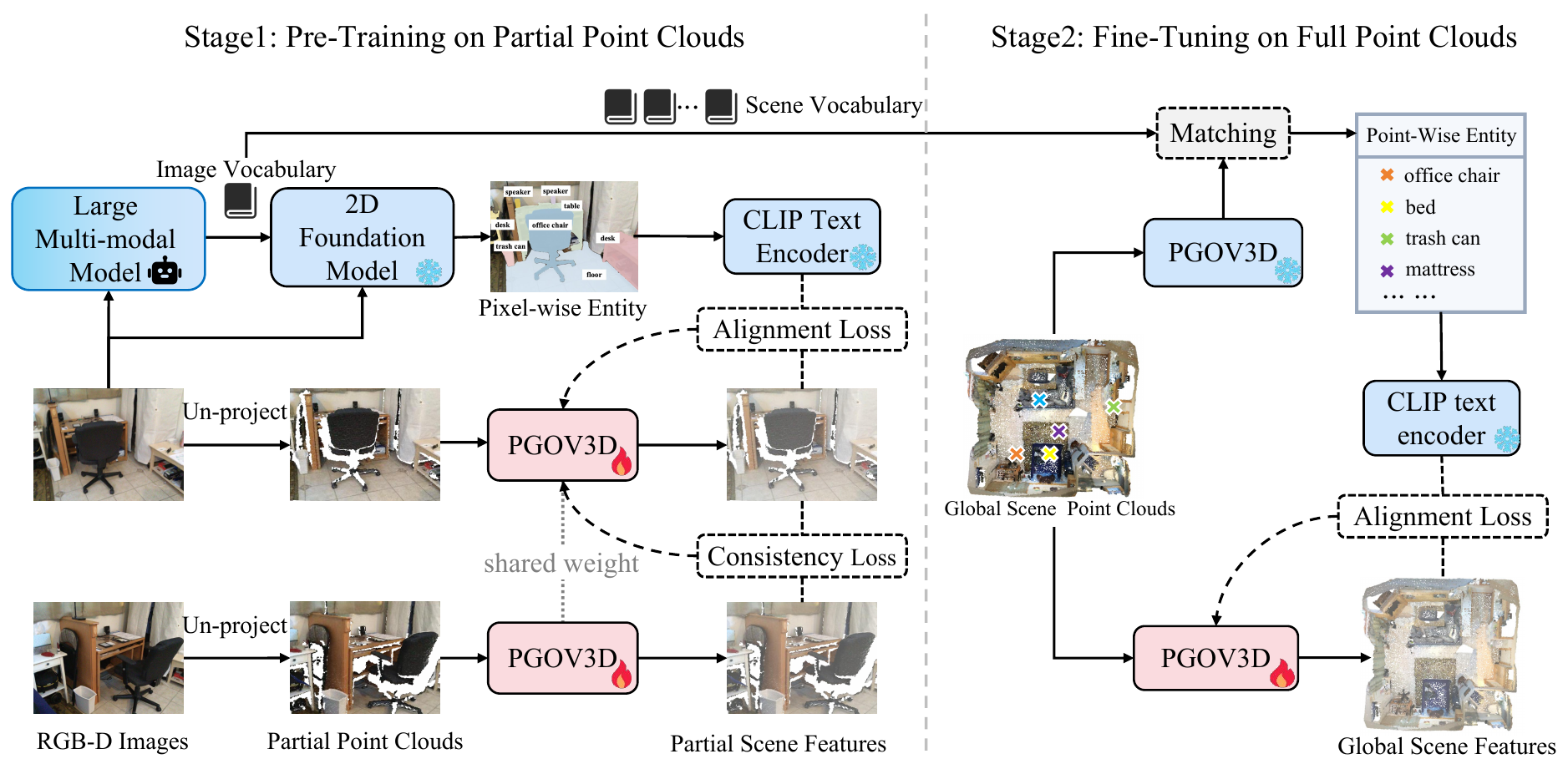} 
    \caption{\textbf{An overview of our \systemname\ pipeline.} Our framework consists of two stages: In the pre-training stage, the 3D segmentation model are pre-trained on partial scenes derived from multi-view RGB-D images via pixel-wise depth projection. The pixel-wise entity is generated by \llavanext~\cite{liu2024llavanext} and 2D foundation model and serve as aligned supervision signals. An auxiliary inter-frame consistency module is introduced to enforce feature consistency. In the fine-tuning stage, the pre-trained 3D segmentation model generates high-confidence point-wise pseudo labels (\emph{i.e.}, point-wise entity), which are used as self-supervised signals to further train the 3D model with the same architecture.}
    \label{fig:framework}
\end{figure*}

In this work, we propose \systemname, a novel framework for open-vocabulary 3D semantic segmentation based on Partial-to-Global curriculum. Figure~\ref{fig:framework} shows the overall pipeline of \systemname.
For each global scene, we denote the 3D point clouds as $\mathbf{P}$ and the associated multi-view RGB-D images as $\{\mathbf{I}_i\}_{i=1,…,M}$. In the pre-training stage, our initial step involves generating image-specific object categories within each image and corresponding object-level semantic segmentation masks. These pixel-wise semantics provide fine-grained object-level information, referred to as pixel-wise entities. The 3D segmentation network is trained on the partial point clouds derived from RGB-D images and supervised by pixel-wise entities to align point features with open-textual representations.
Additional, an auxiliary tasks \consistencymodule\ is introduced during this stage to enforce feature consistency of same physical points in multi views and to enhance spatial understanding. Given that the final goal is to perform segmentation on complete 3D point clouds, we incorporate a fine-tuning stage to bridge the domain gap between partial point clouds derived from RGB-D images and global 3D scene. 
In fine-tuning stage, we aggregate the partial image vocabularies associated with each scene to get the whole scene vocabularies and generate high-confidence point-wise linguistic semantics using the pre-trained 3D segmentation model. The 3D model are then fine-tuned on the global point clouds with the supervision of predicted point-wise semantics to facilate a tighter alignment between point cloud features and the open-textual features.

\subsection{Pre-training on RGB-D Partial Scenes} 
Compared to the abundance of 2D images, 3D data remains significantly scarcer and more difficult to annotate due to its higher dimensionality and structural complexity. For example, the ScanNet dataset offers only 1,513 entire scenes in 3D point clouds but includes 2.5 million RGB-D frames. 
Hence, leveraging rich visual features from 2D images is critical for enabling open-vocabulary understanding in 3D scenarios. To bridge the modality gap and make use of rich semantic information in multi-view images, we convert RGB-D frames into partial point clouds using pixel-wise depth projection and pre-train our 3D segmentation model on these pseudo labels, thereby preserving the semantic content at the pixel level for downstream 3D understanding.

 \begin{figure}[thbp]
    \centering
    \includegraphics[width=1\linewidth]{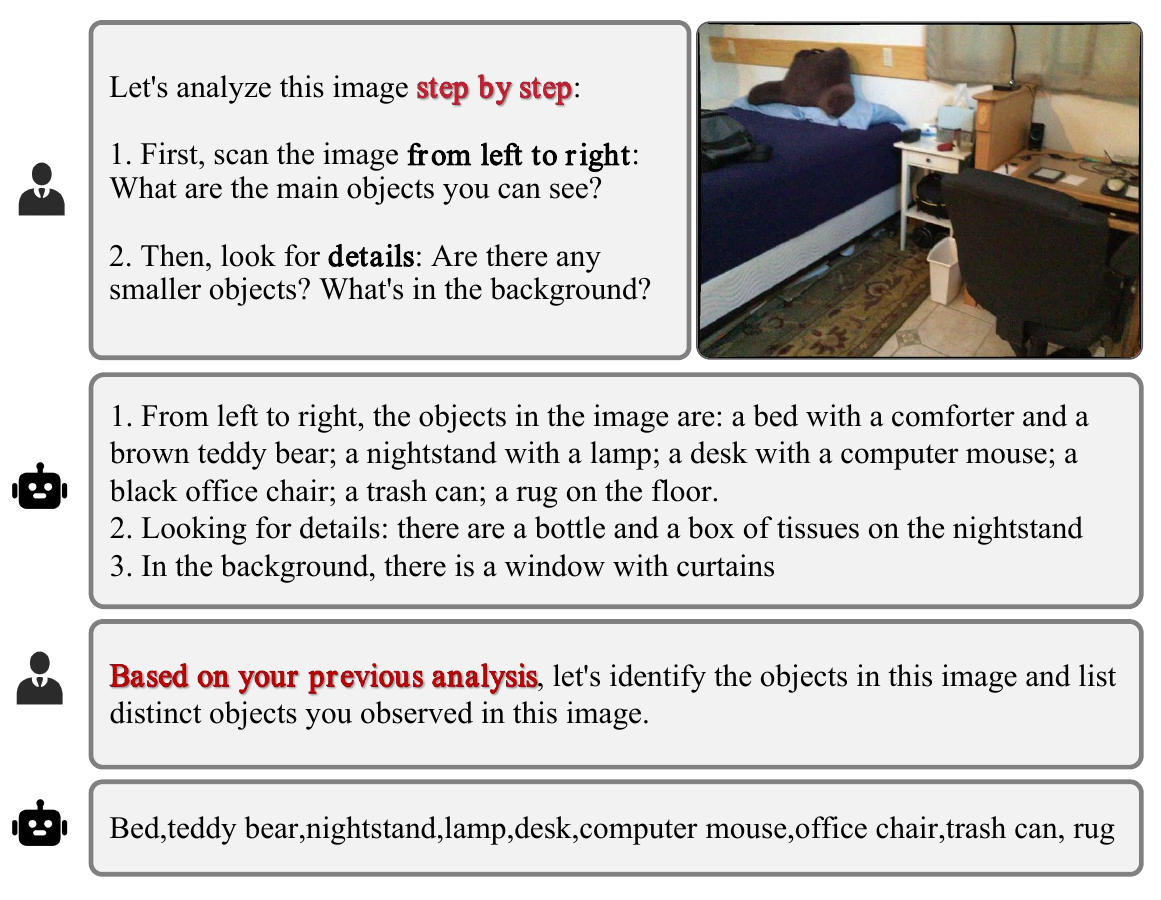} 
    \caption{\textbf{An Illustrative Example of \llavanext~\cite{liu2024llavanext} Interaction.} By prompting the \llavanext~\cite{liu2024llavanext} with Chain-of-Thought ~\cite{wei2022chain} reasoning, we guide it to generate accurate object-centric vocabulary for visual grounding.}
    \label{fig:prompt}
    \vspace{-8pt}
\end{figure}

 \noindent\textbf{Pixel-wise Entity Generation. }
MLMMs demonstrate remarkable open-world visual understanding capabilities, effectively recognizing and describing a wide range of objects beyond closed-set categories. Therefore, we use LLaVA-NeXT~\cite{liu2024llavanext} to identify the object categories $\mathbf{C_i}^{\prime}$ in each image. Rather than directly prompting the model to list object names, we design a Chain-of-Thought (\textit{CoT}) ~\cite{wei2022chain} strategy to guide it toward more accurate and reliable category generation. As shown in Figure~\ref{fig:prompt}, LLaVA-NeXT~\cite{liu2024llavanext} is first prompted to carefully observe and describe the image content, allowing for a more detailed understanding before generating the object list. This staged reasoning process significantly enhances the accuracy and completeness of object identification. Following this, the extracted object list is used to prompt an 2D segmentation foundation model (\emph{e.g}., GroundedSAM~\cite{ren2024grounded}) to grounding object in $\mathbf{C_i}^{\prime}$  and associate successfully grounded categories $\mathbf{C_i}$  with corresponding pixel-level semantic segmentation masks. It enables the establishment of precise pixel-entity correspondences (i.e., pixel-wise entity).

\noindent\textbf{Pixel-wise Depth Mapping to Partial Point Clouds. }
After generating the pixel-wise entities, we convert the images into parse 3D point   clouds $\mathbf{P}_i$ by leveraging the depth information to establish the relationship between points and text. For each RGB-D image $I_i$, given the camera intrinsic matrix $\mathbf{K} \in \mathbb{R}^{3\times 3}$ and pose $\mathbf{T_i} \in \mathbb{R}^{3\times 3}$, each pixel $(u,v)$ with depth $d_{(u,v)}$ is mapped to a 3D point $\mathbf{p}_{(u,v)}$ in world coordinates through a homogeneous transformation:
\begin{equation} 
\mathbf{p_{(u,v)}} 
=
\mathbf{T_i} 
\begin{bmatrix} 
d_{(u,v)} \cdot \mathbf{K}^{-1}\,\bigl[u,\;v\bigr]^{\!\top} 
\end{bmatrix}  .
\label{eq:projection} 
\end{equation} 
A pixel-level partial point scene are obtained by applying Equation \eqref{eq:projection} to all pixels in the RGB-D images. Based on the resulting pixel-wise depth map and entity-pair associations, we can derive point-wise entities by establishing correspondences between the pixel pairs and 3D points. Based on the pixel-to-point correspondence, each 3D point can directly inherit the semantic entity from its corresponding pixel. In this way, the pixel-wise entities are naturally transferred to the point clouds, resulting in point-wise entity annotations for the partial scene.

\noindent\textbf{Vision-Language Alignment on Partial Scenes. } 
 During the pre-training stage, the point-wise entities serve as supervision signals to guide the 3D network in learning visual features aligned with the open-text space. Specifically, we employ a 3D network with a U-Net backbone built upon sparse 3D convolutions to extract the point-wise feature representations $f_i$. For each corresponding entity text, we  select CLIP~\cite{radford2021learning} as the text encoder to obtain open-textual embedding $f_i^t$ because of the open-vocabulary capability and semantic-rich embedding of CLIP. To align the point features with the CLIP text space, we employ the alignment loss that minimizes the cosine distance between the point feature $f_i^p$ and the textual embedding $f_i^t$. We obtain the average alignment loss for the partial scene as:
\begin{equation}
\mathcal{L}_{\text{alignment}} = \frac{1}{N} \sum_{i=1}^{N} \left( 1 - \cos(f_i^p, f_i^t) \right)
\label{eq:alignment_loss}
\end{equation}
This formulation enables the 3D segmentation model to align point-level features with the CLIP feature space, thereby endowing it with the ability to generalize to novel object categories beyond the training set.

\noindent\textbf{Inter-frame consistency module. } 
Since partial RGB-D point clouds are projected from sequential scenes, the same physical points often appear across adjacent frames. Ideally, the feature representations of identical physical points should remain consistent across frames. However, due to lighting variations, occlusions, and viewpoint changes, the model may learn inconsistent features, compromising its performance on full-scene point clouds and leading to unstable semantic segmentation results.

\begin{figure}[thbp]
    \centering
    \includegraphics[width=1.0\linewidth]{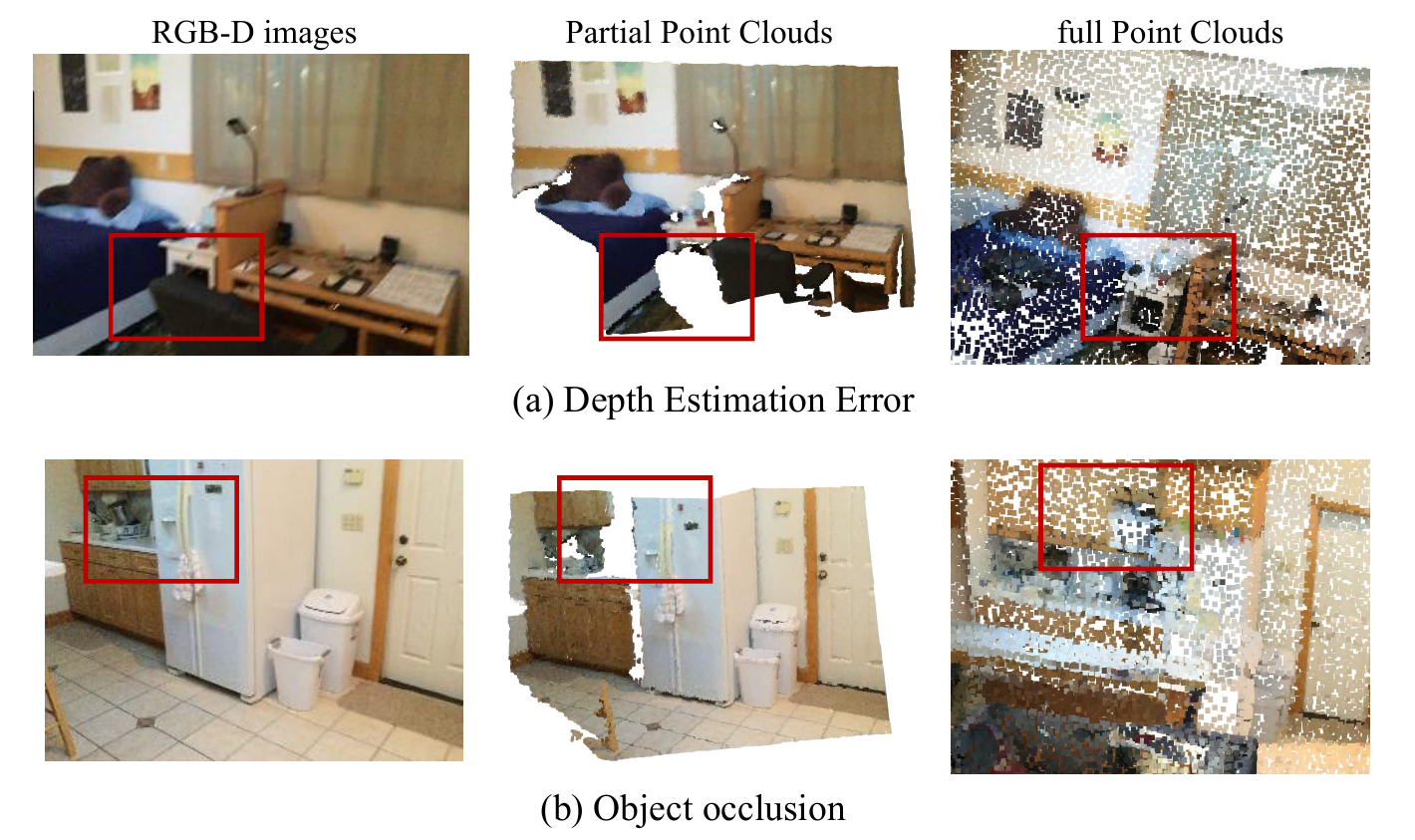} 
    \caption{\textbf{Illustration of domain gap between RGB-D partial point clouds and complete 3D scenes.} (a) demonstrates the discrepancy of depth information. (b) illustrates the object occlusion caused by the limitations of the image viewpoint.}
    \label{fig:gap}
\end{figure}

To address this issue, we introduce an auxiliary module named \consistencymodule, which explicitly enforces feature consistency across overlapping regions under different viewpoints. By encouraging the 3D network to produce consistent features for the same physical points across multiple views, this module mitigates semantic noise caused by conflicting predictions and facilitates the reconstruction of more coherent 3D spatial representations from fragmented observations.

Specifically, let $\mathbf{P}_1$ and $\mathbf{P}_2$ denote the partial point clouds from two consecutive frames, and let $A$ represent the consistent point set comprising matched physical points across these frames. We adopt cosine similarity to measure the directional similarity between feature vectors, which ranges from $[-1, 1]$, with higher values indicating stronger consistency. The cosine similarity between two points is defined as:
\begin{equation}
    S(f_1^p,f_2^p) = \frac{f_1^p \cdot f_2^p}{\Vert f_1^p \Vert \, \Vert f_2^p \Vert},
\label{eq: cosine}
\end{equation}
where $f_1^p$ and $f_2^p$ denote the feature embeddings of corresponding points in $\mathbf{P}_1$ and $\mathbf{P}_2$ respectively.

Based on cosine similarity, we define the 3D consistency loss $\mathcal{L}_{\text{consistency}}$ to enforce alignment between features of matched physical points across frames:
\begin{equation}
\mathcal{L}_{\text{consistency}} = 1 - \frac{1}{|A|} \sum_{(p_1, p_2) \in A} S(f_1^p, f_2^p).
\label{eq:3D loss}
\end{equation}

Therefore, our overall training objective combines both consistency and alignment losses, encouraging stable point-wise features while aligning them with open-set textual semantics:
\begin{equation}
\mathcal{L} = \mathcal{L}_{\text{alignment}} + \lambda \, \mathcal{L}_{\text{consistency}},
\label{eq:total loss}
\end{equation}
where $\lambda$=0.2 is a balancing hyperparameter that controls the trade-off between temporal consistency and semantic alignment.

\subsection{Fine-tuning on Global Scenes} 

Although these partial scenes provide relatively dense local geometry during pre-training, they remain inherently incomplete due to two primary limitations, as illustrated in Figure~\ref{fig:gap}: (a) imprecise depth measurements resulting from surface material properties or color interfering with infrared reflection, and (b) object occlusions arising from changes in camera viewpoints. These issues introduce a domain gap between the RGB-D-based partial point clouds used in pre-training and the complete scene-level point clouds used in the ultimate objective. As a result, directly applying the pre-trained network to full-scene segmentation leads to suboptimal performance.

To mitigate this gap, we introduce a fine-tuning stage that leverages the pre-trained model to automatically generate high-confidence point-wise entity from complete scenes. These pseudo labels are then used to fine-tune the network. This process constructs a seamless and unified vision-language embedding space that is geometrically aligned with full 3D scenes. Notably, the fine-tuning stage requires only unannotated scene-level point clouds and does not rely on any predefined category list.

\noindent \textbf{Point-wise Entity Generation. } 
To enable supervision without manual annotations, we first aggregate the grounded vocabulary $\mathbf{C_i}$ from all RGB-D frames to form a global scene-level vocabulary $\mathbf{C}$. Each entity text in $\mathbf{C}$ is encoded into an open-text embedding using the CLIP text encoder. Simultaneously, we extract 3D point embeddings from the complete scene point clouds using the pre-trained 3D segmentation model. 
Each 3D point in the scene are matched with one entity based on their feature distance within the shared vision-language feature space. 
To enhance the reliability of the point-wise pseudo labels, we adopt a probabilistic smoothing strategy based on repeated grid sampling. Specifically, We first voxelize the point cloud and randomly sample one point from each voxel to form a subset. This sampling process is repeated multiple times to ensure that all points in the scene are adequately covered. 
The pre-trained 3D segmentation model estimates per-point category probability distributions over multiple sampled subsets.
We then average the predicted probabilities of each point and assign the corresponding entity based on the final averaged distribution.
This process provides a scalable mechanism for weak supervision, while mitigating the noise introduced by single-pass predictions or low-confidence associations.

\noindent \textbf{Vision-Language Alignment on Full Scenes. } 
Using the selected reliable point-entity pairs, we fine-tune the 3D segmentation model by optimizing the cosine similarity loss $\mathcal{L}_{consistency}$as defined in Eq.~\ref{eq:3D loss}. In contrast to the pre-training stage, which operates on dense but partial RGB-D views, this fine-tuning phase leverages sparse yet complete scene-level geometry. This shift in supervision granularity enables the model to better capture global spatial context and reinforces the alignment between 3D point features and open-vocabulary textual entities. As a result, the model demonstrates improved generalization in open-vocabulary 3D semantic segmentation tasks.
\section{Experiments}
\subsection{Experiments set up}

\textbf{Datasets. }
We conduct comprehensive experiments on three widely adopted public benchmarks for indoor 3D scene understanding: ScanNet ~\cite{dai2017scannet}, ScanNet200 ~\cite{rozenberszki2022language} and S3DIS ~\cite{7780539}. ScanNet comprises 1,613 RGB-D reconstructed indoor scenes, providing over 2.5 million posed RGB-D images and densely annotated point clouds with 20 standard semantic categories. ScanNet200 increases the level of difficulty by extending the label space to 200 fine-grained object categories. S3DIS consists of 271 room-level point clouds collected from six large-scale indoor areas, covering a variety of spatial layouts and scene types. We restrict our experiments to Area 5 of S3DIS, which is commonly used as the test split.  

\begin{table}[t]
\centering

\caption{\textbf{Performance comparison of several annotation-free 3D semantic segmentation methods on ScanNet and ScanNet200.} Among the reported results, bold indicates the best result; underline denotes the second-best. }
\renewcommand{\arraystretch}{1.3}
\setlength{\tabcolsep}{8.5pt}
\begin{tabular}{l|cc|cc }
\hline
\rowcolor[gray]{.9} & \multicolumn{2}{c|}{\textbf{ScanNet}} & \multicolumn{2}{c}{\textbf{ScanNet200}} \\
\rowcolor[gray]{.9}\multirow{-2}{*}{\textbf{Method}} & mIoU & mAcc & mIoU & mAcc \\
\hline
\hline
Fully-Sup.                             & 72.0 & 80.7 & 23.9 & 32.9 \\
\hline
MSeg Voting~\cite{lambert2020mseg}                         & 45.6 & 54.4 & -    & -    \\
PLA~\cite{ding2023pla}                              & -  &  & 1.8 & -     \\
OpenScene-2D~\cite{peng2023openscene}                    & 50.0 & 62.7 & -    & -    \\
OpenScene-3D~\cite{peng2023openscene}     & 52.9 & 63.2 & 7.3 & - \\
OpenScene~\cite{peng2023openscene}                           & 54.2 & 66.6    & 5.9 & 10.2 \\
RegionPLC~\cite{yang2024regionplc}                         & -  &  &  6.5 & -     \\ 
OV3D~\cite{jiang2024open}                                  & \underline{57.3} & \underline{72.9} & \underline{8.7} & - \\
\hline
\textbf{PGOV3D (Ours)}               & \textbf{59.5} & \textbf{73.2} & \textbf{9.3} & \textbf{17.1} \\
\hline
\end{tabular}
\vspace{-10pt}
\label{tab:comparison}
\end{table}

\noindent\textbf{Evaluation setting and Metrics. }
We first evaluate \systemname\ on ScanNet to assess its zero-shot 3D semantic segmentation performance without annotations or task-specific fine-tuning. To further test its generalization and open-vocabulary capabilities in more challenging settings, we also conduct experiments on ScanNet200, which features a larger and more fine-grained label space.
We further evaluate \systemname\ under a supervised setting using annotated data for a comprehensive comparison across learning paradigms. For quantitative analysis, we report mean Intersection-over-Union (mIoU) and mean Accuracy (mAcc).

\noindent\textbf{Implementation Details. }
In the process of generating pixel-wise entities, we use \llavanext ~\cite{liu2024llavanext} as the MLMM to identify object vocabulary in the image and Grounded-SAM~\cite{ren2024grounded} as the 2D foundation model to generate 2D semantic masks. For vision-language alignment, point features are extracted with SparseConvNet ~\cite{Choy_Gwak_Savarese_2019} and text features with CLIP ~\cite{radford2021learning}. Pre-training uses ~25k frames sampled every 100 frames from RGB-D sequences, while fine-tuning is conducted on 1,201 reconstructed indoor scans. The model is trained with AdamW ~\cite{Loshchilov_Hutter_2017}, using a batch size of 30 for pre-training and 6 for fine-tuning.

\begin{table*}[th]
    \centering
    \caption{\textbf{Performance Comparison of different methods for base-annotated 3D semantic segmentation on ScanNet.} mIoU$^B$ indicates the mIoU of base categories, and mIoU$^N$ indicates the mIoU of novel categories. Bold indicates best performance, and underline indicates second best.}
    \renewcommand{\arraystretch}{1.2}
    \setlength{\tabcolsep}{10pt}
    \begin{tabular}{lccc|ccc|ccc}
        \hline
        \rowcolor[gray]{.92}
         & 
        \multicolumn{3}{c|}{\textbf{B10/N9}} & 
        \multicolumn{3}{c|}{\textbf{B12/N7}} & 
        \multicolumn{3}{c}{\textbf{B15/N4}} \\
        \rowcolor[gray]{.92}
        \multirow{-2}{*}{\textbf{Method}} & hIoU & mIoU$^B$ & mIoU$^N$ 
        & hIoU & mIoU$^B$ & mIoU$^N$ 
        & hIoU & mIoU$^B$ & mIoU$^N$ \\
        \hline
        \hline
        Fully-Sup. & 71.5 & 77.0 & 66.7 & 72.1 & 72.2 & 72.0 & 74.6 & 70.2 & 79.6\\ 
        \hline
        3DGenZ~\cite{michele2021generative} & 12.0 & 63.6 & 06.6 & 19.8 & 35.5 & 13.3 & 20.6 & 56.0 & 12.6 \\ 
        3DTZSL~\cite{cheraghian2020transductive}& 07.8 & 55.5 & 04.2 & 03.8 & 36.6 & 02.0 & 10.5 & 36.7 & 06.1 \\
        LSeg-3D~\cite{guo2019seg} & 01.8 & 68.4 & 00.9 & 00.9 & 55.7 & 00.1 & 00.0 & 64.4 & 00.0 \\
        PLA~\cite{ding2023pla} & 53.1 & 76.2 & 40.8 & 55.3 & 69.5 & 45.9 & 65.3 & 68.3 & 62.4 \\
        RegionPLC~\cite{yang2024regionplc} & \underline{58.8} & \textbf{76.6} & \underline{47.7} & \underline{65.1} & 69.6 & \underline{61.1} & \underline{69.9} & \underline{68.4} & \textbf{71.5}  \\
        XMask3D~\cite{wang2024xmask3d} &55.7 &\underline{76.5} &43.8 &61.7 &\underline{70.2}& 55.1 & \textbf{70.0} & \textbf{69.8} & \underline{70.2}  \\
        \hline
        \systemname~(ours) & \textbf{65.3} & 75.8 & \textbf{57.3} & \textbf{68.1} & \textbf{70.3}& \textbf{66.0} & 68.6 & \underline{68.4} & 69.3\\
        
        \bottomrule
    \end{tabular}
    \label{tab:bn}
\end{table*}

\begin{figure*}[thbp]
    \centering
    \includegraphics[width=1.0\linewidth]{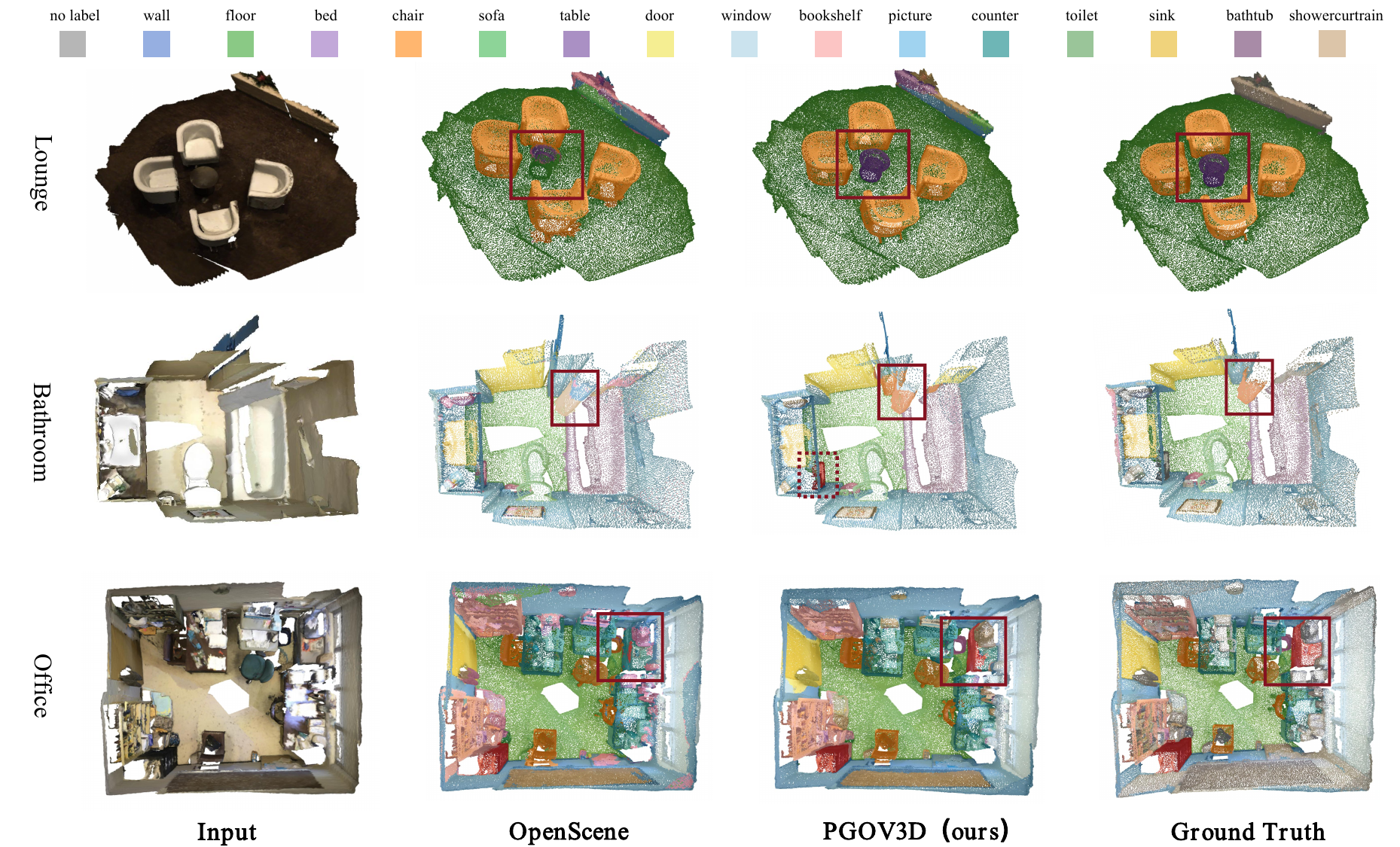} 
    \caption{\textbf{Qualitative segmentation comparison between \systemname\ and OpenScene on the ScanNet dataset.} The regions highlighted with red boxes represent the same areas segmented using different methods, while the areas enclosed by red dashed boxes indicate regions that were not labeled in the Ground Truth but were correctly identified by \systemname.}
    \label{fig:compare}
\end{figure*}

\subsection{Main Results}
\textbf{Comparison on Zero-Shot 3D semantic Segmentation. } 
We evaluate our method on ScanNet and compare it with several annotation-free and zero-shot 3D semantic segmentation methods. As shown in Table~\ref{tab:comparison} (left)
, \systemname\ achieves 59.5 mIoU and 73.2 mAcc, outperforming all existing annotation-free approaches. Notably, compared with the most competitive baseline, OV3D, our method achieves a +2.2 mIoU improvement. These results underscore its effectiveness in bridging 2D vision-language models and 3D semantic understanding. Although fully-supervised methods still lead by a notable margin (\emph{e.g.}, 72.0 mIoU vs. 59.5), our approach significantly narrows this gap without requiring manual annotations, demonstrating strong potential for real-world deployment.


\noindent\textbf{Adaptability in Long-tailed Scenarios. }
To evaluate the fine-grained generalization ability of \systemname\ in large-scale open vocabulary settings, we further conduct experiments on the ScanNet200 dataset. Unlike the original ScanNet benchmark with 20 categories, ScanNet200 provides 200 fine-grained semantic categories, significantly increasing scene complexity and semantic diversity. This setting imposes a greater challenge for segmentation models due to increased intra-class variance and inter-class similarity. As shown in Table~\ref{tab:comparison} (right), \systemname\ achieves an mIoU of 9.3 and an mAcc of 17.1, outperforming baseline methods such as OV3D (8.7). Although there is still a notable performance gap compared to the fully-supervised upper bound (21.3), our results highlight the effectiveness of the proposed Point-Entity alignment strategy in handling large category shifts and promoting robust knowledge transfer in annotation-free settings.

\noindent\textbf{Base and Novel Category Setting. }
In this section, we evaluate the capability of \systemname\ to recognize novel categories under the open-world base-annotated setting on the ScanNet. We follow prior works to divide the 19 semantic classes from ScanNet (excluding other furniture) into two disjoint subsets: base categories with annotations (B) and novel categories without any annotation (N). We adopt three standard splits B15/N4, B12/N7, and B10/N9, to comprehensively assess the performance across varying levels of supervision. Since \systemname\ is inherently annotation-free, we simulate base supervision by replacing the point-level entities of base categories with their corresponding ground-truth textual labels, while keeping the entities for novel categories unchanged.  This design allows us to explore the model’s ability to generalize to unseen novel categories without requiring any additional supervision. As shown in Table~\ref{tab:bn}, \systemname\ consistently achieves competitive performance across all splits. 
In the most challenging B10/N9 split, \systemname\ achieves a harmonic IoU (hIoU) of 65.3, significantly outperforming all baseline methods. Notably, it attains a novel category mIoU of 57.3, surpassing the second-best model (RegionPLC, 47.7) by a substantial margin of +9.6\%, highlighting its superior ability to recognize unseen categories with minimal supervision. On the B12/N7 split, \systemname\ maintains its leadership, outperforming all other approaches once again.
We observe that \systemname\ performs better when there are more novel categories. We attribute this to the increased category diversity that comes with a larger number of novel classes, which encourages \systemname\ to learn a broader and more diverse set of concepts. This, in turn, facilitates better alignment with the open-ended nature of language supervision, demonstrating the strong generalization ability of \systemname\ in complex and open environments.

\noindent\textbf{Knowledge transformer among Datasets. }
To assess the adaptability of \systemname\ under long-tailed distributions, we conduct experiments on the S3DIS dataset in a zero-shot domain transfer setting. Specifically, the model is trained solely on the ScanNet dataset without any annotations, and directly evaluated on S3DIS using mIoU as the metric. This setting is particularly challenging due to the domain gap and class imbalance inherent in indoor scenes. As shown in Table~\ref{tab:s3dis}, \systemname\ achieves an mIoU of 43.2, outperforming several recent methods such as OV3D (41.3) and RegionPLC (36.9). These results demonstrate the model's strong ability to generalize to unseen domains and adapt to long-tailed categories, even in the absence of manual annotations.


\begin{table}[th]
    \centering
    \caption{\textbf{Performance comparison for zero-shot domain transfer from ScanNet to S3DIS.} The model is trained on ScanNet (annotation-free) and evaluated on S3DIS using mIoU.}
    \renewcommand{\arraystretch}{1.2}
    \begin{tabular}{l|c c c c c}
        \hline
        \cellcolor[gray]{.92}\textbf{Method} & PLA~\cite{ding2023pla} & RegionPLC~\cite{yang2024regionplc} & OV3D~\cite{jiang2024open} & \systemname \\
        \hline
        \cellcolor[gray]{.92}\textbf{mIoU} & 13.4 & 36.9 & 41.3 & 43.2 \\
        \hline
    \end{tabular}
    \label{tab:s3dis}
\end{table}


\noindent\textbf{Qualitative Results.} Fig.\ref{fig:compare} shows qualitative comparisons on the ScanNet benchmark across three representative indoor scenes: office, lounge, and bathroom. These examples are chosen to reflect diverse spatial layouts and semantic complexities commonly encountered in indoor environments. Overall, \systemname\  produces segmentation results that are more consistent with the ground truth compared to existing approaches, particularly in terms of boundary precision and object-level accuracy. For instance, as highlighted by the red boxes in Fig.\ref{fig:compare}, our model successfully segments the cabinet in the office scene, a structure missed by other methods. In the lounge scene, \systemname\ achieves sharper and more coherent boundaries around the table, demonstrating enhanced capability in capturing fine-grained geometric details.

\subsection{Ablation Study}
\begin{figure}[thbp]
    \centering
    \includegraphics[width=1.0\linewidth]{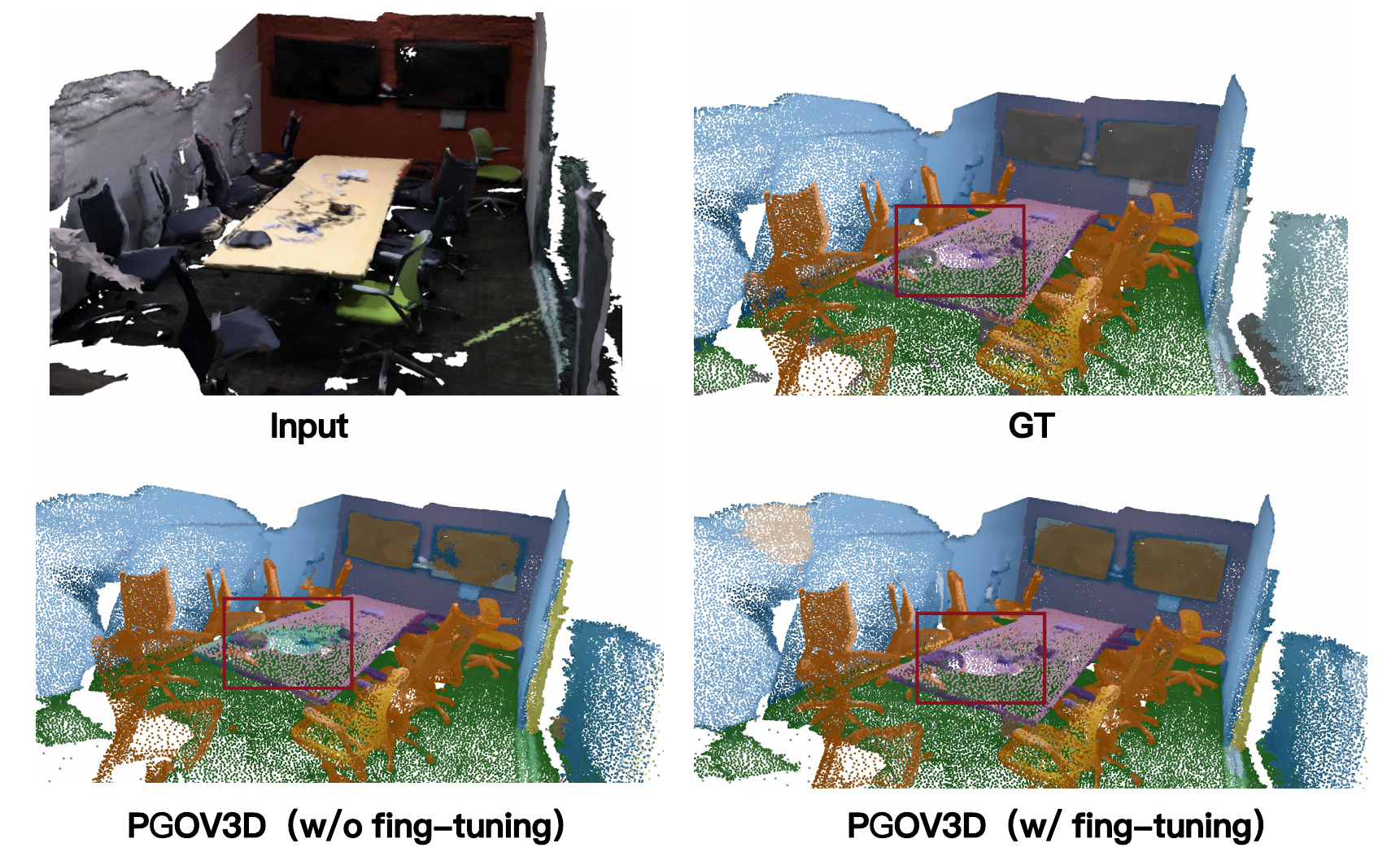} 
    \caption{\textbf{Qualitative segmentation with and without further fine-tuning module.} The regions enclosed by red solid lines in the figure represent the same areas within the same scene that have been segmented by different methods.}
    \label{fig:stage-ablation}
\end{figure}

\noindent\textbf{Effect of Two-Stage Training Strategy. }
We compare the performance of \systemname\ solely pre-trained on images and further fine-tuned on global point clouds. As shown in
table \ref{tab:stage-ablation} the incorporation of the fine-tuning stage leads to an improvement of 4.77 in mIoU and 4.24 in mAcc. The qualitative comparison in Fig.\ref{fig:stage-ablation} also highlights the benefit of fine-tuning. While \systemname\ (without stage 2) yields partially inaccurate segmentation, especially around object boundaries, the fine-tuned version demonstrates significantly more accurate and complete object delineation. This experiment confirms the effectiveness of Partial-to-Global framework.

\begin{table}[h]
    \centering
     \caption{\textbf{Performance with and without further fine-tuning module.} \textbf{Stage 1} refers to the prediction of the entire scene segmentation using the model trained in the first stage only, \textbf{Stage 2} indicates the use of a model that is trained in first stage and further fine-tuned in the second stage.}
     \renewcommand{\arraystretch}{1.2}
    \setlength{\tabcolsep}{13pt}
    \begin{tabular}{c c|c|c}
        \hline
        \rowcolor[gray]{.92}
        \textbf{Stage 1} & \textbf{Stage 2} & \textbf{mIoU} & \textbf{mAcc} \\
        \hline
        \hline
        \ding{51} & \ding{55} & 54.7&68.1  \\  
        \ding{51} & \ding{51} & \textbf{59.5}\textcolor{green}{(+4.8)}&\textbf{73.2}\textcolor{green}{(+5.1)}   \\ 
        \hline
    \end{tabular}
    \label{tab:stage-ablation}
\end{table}

\noindent\textbf{Impact of Inter-frame Consistency Module.}
We further analyze the impact of \consistencymodule\ on model performance. Specifically, we ablate the \consistencymodule\ from the first stage and train the network using only the primary loss function $\mathcal{L}_{alignment}$ as in Eq. \ref{eq:alignment_loss}. As shown in Table ~\ref{tab:consistency-ablation}, the absence of the consistency module results in a 1.31\% drop in mIoU during the first stage, which subsequently affects the overall performance in the second stage. These results demonstrate the effectiveness of the consistency module in enhancing feature alignment and improving open-vocabulary semantic segmentation performance.

\begin{table}[h]
    \centering
    \caption{\textbf{Performance with and without the consistency module.} Here, 'w' denotes the use of the consistency module, while 'w/o' indicates its absence.}
    \setlength{\tabcolsep}{12pt}
    \renewcommand{\arraystretch}{1.2}
    \begin{tabular}{l|c c}
        \hline
        \rowcolor[gray]{.92}
        \textbf{Method} & \textbf{stage 1 mIoU} &  \textbf{stage 2 mIoU}\\
        \hline
        \hline
        w/o consistency & 53.3& 57.2\\ 
        w consistency & \textbf{54.7}\textcolor{green}{(+1.4)} & \textbf{59.5}\textcolor{green}{(+2.3)}\\
        \hline
    \end{tabular}
    \vspace{11pt}
    \label{tab:consistency-ablation}
\end{table}

\noindent\textbf{Benefit of Loading Pre-weight in stage 2. } As shown in Table \ref{tab:pre-train-ablation}, we investigate the influence of loading pre-trained weights during fine-tuning. Compared to training from scratch, initializing the model with pre-trained weights leads to consistent improvements in both mIoU (+0.7) and mAcc (+1.4). These results highlight the importance of leveraging prior knowledge from partial point clouds, which helps the model better adapt to the global scene context and accelerates convergence during the fine-tuning process.

\begin{table}[h]
    \centering
    \caption{\textbf{Performance with and without the consistency module.} Here, 'w' denotes the use of the consistency module, while 'w/o' indicates its absence.}
    \setlength{\tabcolsep}{9pt}
    \renewcommand{\arraystretch}{1.2}
    \begin{tabular}{l|c c}
        \hline
        \rowcolor[gray]{.92}
        \textbf{Method} & \textbf{mIoU} &  \textbf{mACC}\\
        \hline
        \hline
        stage2 w/o pre-trained weight & 58.8 & 71.8\\ 
        stage2 w/ pre-trained weight & \textbf{59.5}\textcolor{green}{(+0.7)} & \textbf{73.2}\textcolor{green}{(+1.4)}\\
        \hline
    \end{tabular}
    \label{tab:pre-train-ablation}
\end{table}
\section{Conclusion}
In this work, we introduce \systemname, a two-stage training paradigm that improves open-vocabulary 3D semantic segmentation with partial-to-global curriculum.
The core idea of \systemname\ is to leverage dense partial-view supervision guided by large vision-language models in the first stage, and then adapt to full-scene understanding via pseudo-labeled fine-tuning. Benefiting from this Partial-to-Global curriculum and the introduced consistency module, \systemname\  achieves fine-grained, generalizable 3D understanding across diverse open-vocabulary categories. 
While \systemname\ makes a step forward in scalable open-world 3D perception, challenges remain in handling dynamic environments and long-term adaptation. In future work, we plan to extend this framework to large-scale, real-time scenarios and explore its integration with continual learning paradigms for sustained deployment in open-world settings.

\clearpage
\bibliographystyle{IEEEtran}
\bibliography{IEEEabrv, reference}

\end{document}